\documentclass[times, review, 10pt]{elsarticle}

\usepackage{xcolor}
\usepackage{multirow}
\usepackage{makecell}
\usepackage{stmaryrd}
\usepackage{booktabs}

\usepackage{amssymb}
\usepackage{amsmath}
\usepackage{setspace}
\usepackage{array}
\usepackage{tabularx}
\usepackage{calc}
\journal{ }

\begin{document}

\begin{frontmatter}



\title{ Multi-Modal Face Anti-Spoofing via Cross-Modal Feature Transitions} 



\author[2,fn1]{Jun-Xiong Chong}
\ead{jxchong@gapp.nthu.edu.tw}
  
\author[2]{Fang-Yu Hsu}
\ead{shellyhsu@gapp.nthu.edu.tw}

\author[2]{Ming-Tsung Hsu}
\ead{xmc510063@gapp.nthu.edu.tw}

\author[2]{Yi-Ting Lin}
\ead{yitinglin@gapp.nthu.edu.tw}

\author[2]{Kai-Heng Chien}
\ead{ss113062582@gapp.nthu.edu.tw}

\author[2]{Chiou-Ting Hsu}
\ead{cthsu@cs.nthu.edu.tw}

\author[1,2]{Pei-Kai Huang\corref{cor1}}
\ead{alwayswithme@gapp.nthu.edu.tw}

\cortext[cor1]{Corresponding author}
\fntext[fn1]{This is the first author.}

\affiliation[1]{organization={College of Computer and Cyber Security,  Fujian Normal University},
                city={Fuzhou},
                country={China}
                }

\affiliation[2]{organization={Department of Computer Science, 
                              National Tsing Hua University},
                city={Hsinchu},
                country={Taiwan}}

\begin{abstract} 
Multi-modal face anti-spoofing (FAS) aims to detect genuine human presence by extracting discriminative liveness cues from multiple modalities, such as RGB, infrared (IR), and depth images, to enhance the robustness of biometric authentication systems.
However, because data from different modalities are typically captured by various camera sensors and under diverse environmental conditions, multi-modal FAS often exhibits significantly greater distribution discrepancies across training and testing domains compared to single-modal FAS.
Furthermore, during the inference stage, multi-modal FAS confronts even greater challenges when one or more modalities are unavailable or inaccessible.
In this paper, we propose a novel Cross-modal Transition-guided Network (CTNet) to tackle the challenges in the multi-modal FAS task.
Our motivation stems from that, within a single modality, the visual differences between live faces are typically much smaller than those of spoof faces. Additionally, feature transitions across modalities are more consistent for the live class compared to those between live and spoof classes.
Upon this insight, we first propose learning consistent cross-modal feature transitions among live samples to construct a generalized feature space.
Next, we introduce learning the inconsistent cross-modal feature transitions between live and spoof samples to effectively detect out-of-distribution (OOD) attacks during inference.
To further address the issue of missing modalities, we propose learning complementary infrared (IR) and depth features from the RGB modality as auxiliary modalities.
Extensive experiments demonstrate that the proposed CTNet outperforms previous two-class multi-modal FAS methods across most protocols.  

\end{abstract}




\end{frontmatter}

\section{Introduction}
 
While facial recognition conveniently enables access to various platforms, devices, and payment systems, it also confronts increasing threats from face spoofing attacks, including 3D mask attacks (i.e., wearing a face mask), print attacks (i.e., printing a face on paper), and replay attacks (i.e., displaying a face video).
To tackle facial spoof attacks, many single-modal Face Anti-Spoofing (FAS) methods have been developed using the most common modality, i.e., RGB images, 
to distinguish between live and spoof faces. 
Some recent methods \cite{huang2024one,huang2025slip,huang2025unsupervised} focused on learning discriminative features exclusively from live training images.
Other methods \cite{jia2020single,huang2022learnable}, noting that live images collected by RGB sensors exhibit smaller distribution discrepancies compared to spoof images, proposed aggregating all real faces while separate fake ones from different training domains to learn a generalized feature space.
However, as new spoofing attacks evolve to mimic the intricate textural details of live faces, FAS methods relying on a single modality become increasingly insufficient to counter these rapidly advancing threats.
Therefore, recent multi-modal FAS methods, which incorporate additional modalities such as infrared (IR) and depth images, have emerged as a favorite to better capture discriminative liveness information in face anti-spoofing.

\begin{figure}[!ht]
    \centering
    \includegraphics[width=0.75\linewidth]{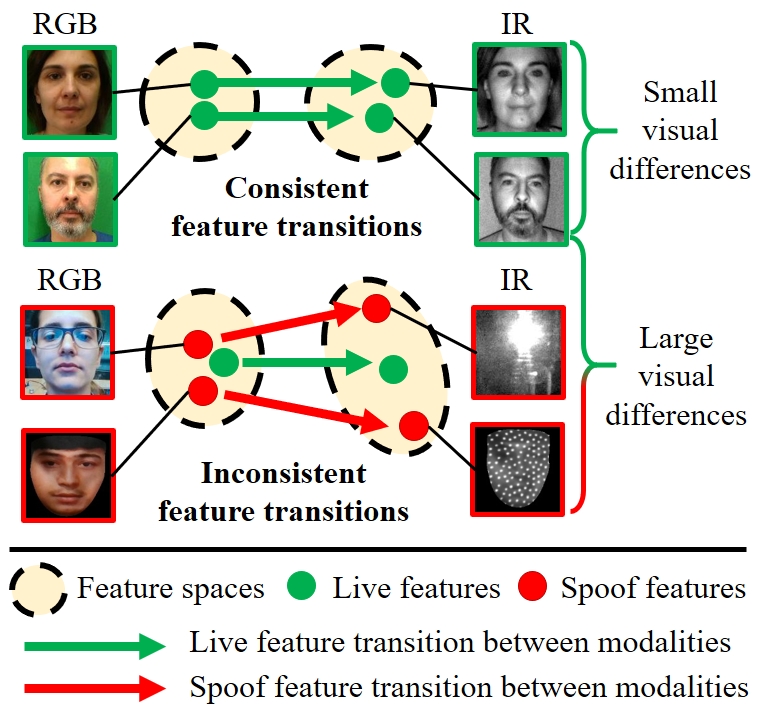} 
\caption{  
Illustration of the proposed multi-modal face anti-spoofing (FAS) method via cross-modal feature transitions.
Within a single modality, live images (green boxes) typically exhibit smaller visual differences, whereas the difference between live (green boxes) and spoof images (red boxes) is larger. As a result, feature transitions across modalities are more consistent for live images compared to those between live and spoof images.
We leverage these properties and focus on learning consistent feature transitions across modalities for live images and inconsistent feature transitions between live and spoof images. 
} 
\label{fig:idea}   
\end{figure}  
  
\begin{figure}  
    \centering
     \includegraphics[width=0.9\columnwidth]{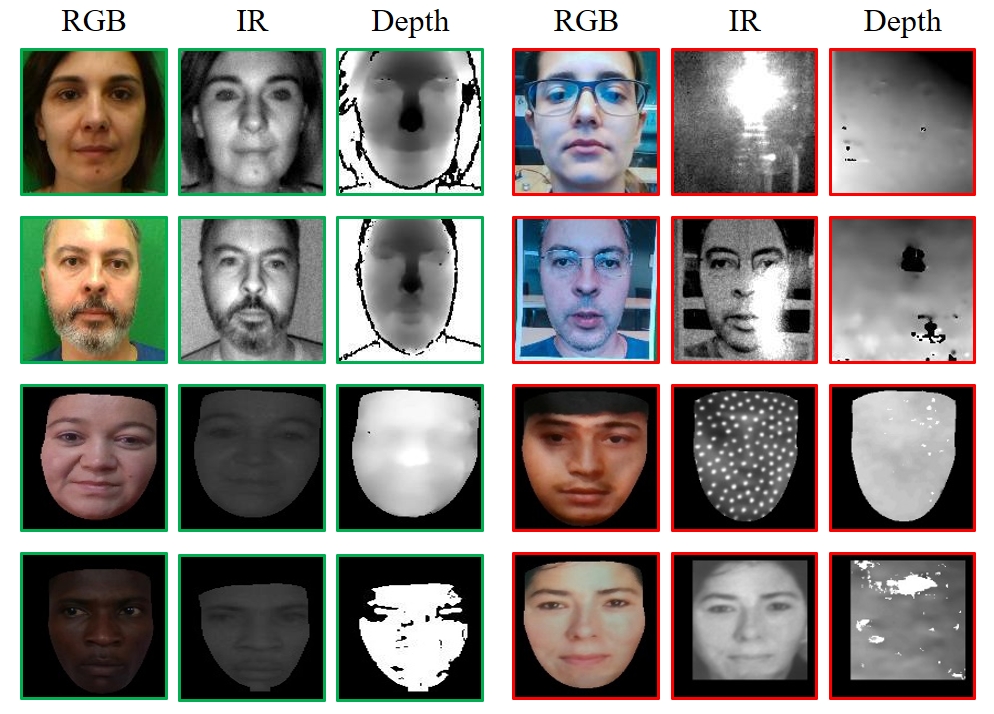}
\caption{  Examples of live images (green boxes) and spoof images (red boxes) in RGB, IR, and depth modalities.} 
\label{fig: different modalities}  
\end{figure}

Most multi-modal FAS methods \cite{li2023learning,deng2023attention,liu2023fm,yu2020multi} focused on fusing different modality-specific features to construct a more discriminative latent feature representation.
For example, in \cite{george2021cross}, the authors proposed to weight the contributions of different modalities for fusing features to enhance multi-modal FAS.
Similarly, the authors in \cite{liu2023fm} proposed utilizing a cross-attention mechanism to fuse features extracted from different modalities.
While multiple modalities provide richer information than a single modality for detecting spoof attacks, multi-modal FAS still meets several challenges.
The first challenge lies in the distribution discrepancies between training and testing domains across different modalities. For example, as shown in Figure~\ref{fig: different modalities}, IR and depth modalities exhibit more diverse visual differences between the live and spoof classes compared to the RGB modality.
Moreover, within the same modality, different spoof attacks can also produce substantial visual differences. Therefore, learning a generalized feature space from modalities with such significant visual variations becomes particularly challenging for multi-modal FAS.
The next challenge concerns detecting previously unseen spoof attacks. Specifically, a FAS model that performs well on specific training datasets may fail to detect attacks not included in the training data and thus results in performance degradation during inference in real-world scenarios.
Finally, since IR and depth modalities are not as readily available as RGB, multi-modal FAS often encounters the issue of missing modalities (e.g., IR or depth images) during inference \cite{yu2023flexible}.

Our motivation stems from the observation that, within a single modality, the visual differences among live faces are typically much smaller than those among spoof faces. As noted in \cite{jia2020single}, RGB images of the live class exhibit smaller distribution discrepancies compared to those of the spoof class. 
Similarly, we observe that the IR and depth modalities of the live class have similar characteristics within each modality.
To verify this, in Section~\ref{sec: similar_live  feature}, we conduct a correlation analysis across the entire multi-modal FAS dataset \textbf{CeFA} to investigate the within-class feature similarity for each modality. As shown  
in Figure~\ref{fig:single}, all three single-modal features of the live class demonstrate high similarity and exhibit smaller visual differences compared to those of the spoof class.
Next, inspired by \cite{sun2023rethinking}, which proposed aligning the live-to-spoof transition (i.e., the transition from live to spoof samples) to construct a domain-invariant live-vs-spoof hyperplane for learning generalized liveness features, we suspect that feature transitions between different modalities may exhibit distinct characteristics for live and spoof classes.
To investigate this, in Section~\ref{sec:feature transitions}, we conduct a correlation analysis on feature transitions across different modalities to examine the differences between live and spoof classes.
As shown in Figure~\ref{fig:transition}, the feature transitions across modalities (e.g., RGB to IR, RGB to Depth, and IR to depth) within the live class demonstrate relatively higher correlation compared to those between the live and spoof classes.
These findings, which will be further discussed in Section~\ref{sec:Characteristics}, suggest that consistent feature transitions across modalities within the live class and inconsistent transitions between the live and spoof classes may serve as key distinguishing characteristics of multi-modal FAS. 
   
In this paper, we build on the two aforementioned findings—consistent feature transitions across modalities within the live class and inconsistent feature transitions between the live and spoof classes—and propose a novel feature learning approach to address the challenges of multi-modal FAS. 
Figure \ref{fig:idea} illustrates the proposed idea.
Building on the first finding, we propose learning a generalized feature space by aligning feature transitions across modalities using only live images. Expanding on the second finding, we further incorporate spoof images to push the cross-modal feature transitions of the spoof class away from those of the live class to enhance the discriminability of latent features and to improve detecting out-of-distribution (OOD) attacks during inference.
In addition, to address the issue of missing modalities, we introduce an effective complementary feature learning approach to derive IR-like and depth-like features from the RGB modality as auxiliary modalities.
Our experimental results from both intra-domain and cross-domain tests show that the proposed method outperforms previous multi-modal FAS techniques on the majority of protocols.

Our contributions are summarized as follows:
   
\begin{itemize}
\item 
We conduct a correlation analysis on feature transitions across different modalities using the multi-modal FAS dataset \textbf{WMCA}, and observe two key findings: (1) consistent feature transitions across modalities within the live class, and (2) inconsistent feature transitions between the live and spoof classes.

\item   
We propose a novel feature learning approach based on cross-modal feature transitions for multi-modal face anti-spoofing. In particular, we introduce learning a generalized and discriminative feature space by leveraging two key findings: consistent feature transitions within the live class and inconsistent feature transitions  between live and spoof images.   

\item  
To address the challenge of missing modalities during inference, we further propose an effective complementary feature learning approach to extract IR-like and depth-like features from RGB images as auxiliary modalities.

\item   
Extensive experiments demonstrate that the proposed method outperforms previous multi-modal FAS methods on most of the protocols.  

\end{itemize}

\section{Related Work}
\label{sec:Related Work}

\subsection{Single-Modal Face Anti-Spoofing}  
Many single-modal face anti-spoofing (FAS) methods have been developed to counter potential spoofing attacks and focused on learning discriminative liveness features and addressing cross-domain challenges.
In \cite{jia2020single}, the authors leveraged the characteristics of live data to learn a feature space where live features are compactly clustered, while ensuring that spoof features are dispersed yet compact across domains. Similarly, the authors in \cite{huang2022learnable,liao2023domain} proposed clustering both live and spoof features, with the spoof features being grouped based on their attack types.
Next, in \cite{huang2023test}, the authors suggested adapting FAS models to target domains by clustering live features towards a global live prototype, while spoof features are clustered towards a local spoof prototype. 
In \cite{huang2025channel,huanga2025enhancing}, the authors proposed to learn long-range and discriminative features for FAS.

These studies demonstrate that learning liveness information from the live class is indeed possible to detect previously unseen spoof attacks. 
 
Other methods have focused on suppressing domain-specific information or extracting domain-agnostic liveness features. In \cite{wang2022domain}, the authors employed contrastive learning to highlight liveness-related style information while suppressing domain-specific details. In \cite{sun2023rethinking}, the authors addressed domain gaps by encouraging domain separability and aligning the live-to-spoof transition through invariant risk minimization.
The approach in \cite{zhou2023instance} focused on aligning features at the instance level to eliminate the need for coarse domain labels and on incorporating techniques to manage instance-specific styles. In \cite{srivatsan2023flip}, the authors utilized vision transformers with multi-modal pre-trained weights to align image representations with natural language descriptions through a multi-modal contrastive learning strategy for enhancing domain generalizability.
In addition, the authors in \cite{shao2022federated} proposed focusing on feature disentanglement to further improve generalization. 

\subsection{Multi-Modal Face Anti-Spoofing} 
  
With the rise of multi-modality cameras and the growing need for enhanced security in face recognition systems, multi-modal FAS methods have ganthed increasing interest. In \cite{yu2020multi}, the authors proposed a multi-modal extension of CDCN \cite{yu2020searching} to effectively learn fine-grained information from RGB, depth, and infrared modalities.
The authors in \cite{wang2022conv} introduced a novel architecture to seamlessly combines local patch convolution with a global multi-layer perceptron. This design enables extracting both local details and long-range dependencies, which are crucial for face anti-spoofing.
In \cite{li2023learning}, an approach was proposed to leverage RGB and depth inputs by employing an adversarial learning mechanism and an attention-based fusion module for stage-wise modality fusion. The authors in \cite{deng2023attention} addressed challenges related to low-light conditions and focused on enhancing the representation capacity of fused features.
Moreover, the authors in \cite{liu2023fm} relied on fused multi-modal information to derive modality-agnostic representation. In \cite{liu2023ma}, the authors adopted a modified transformer attention module to mitigate the impact of modality-specific features and incorporated cross-modal attention to effectively extract modality-agnostic features. 

\begin{figure}  
    \centering
    \includegraphics[width=0.9\columnwidth]{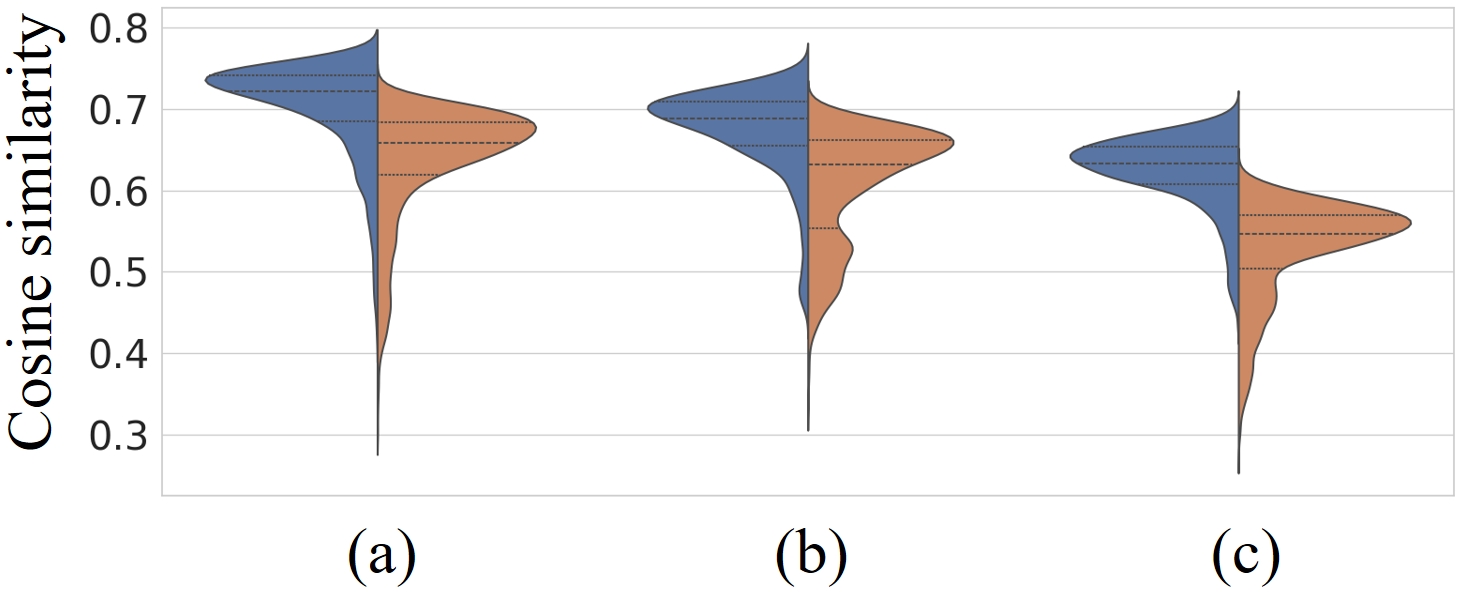}
\caption{   
Distributions of cosine similarity within individual modalities, including (a) RGB images, (b) IR images, and (c) depth images, for the live class (blue) and the spoof class (orange) in the \textbf{CeFA} dataset \cite{liu2021casia}. 
We observe that the spoof class within the same modality exhibits significantly larger visual differences compared to the live class within the same modality.
} 
\label{fig:single} 
\end{figure}

\section{ 
Modal Characteristics in Multi-Modal Face Anti-Spoofing} 
\label{sec:Characteristics}  
In this section, we explore the modal characteristics, particularly focusing on the latent features within individual modalities and the feature transitions between different modalities in multi-modal FAS.

\subsection{  Similar Features within Individual Modalities for the Live Class} 
\label{sec: similar_live  feature}
 
Figure \ref{fig: different modalities} presents examples of live and spoof classes across different modalities from the multi-modal FAS dataset \textbf{WMCA} \cite{george2019biometric}. 
These examples demonstrate that the live class exhibits highly similar visual characteristics within each modality and that each modality possesses its own distinct properties.
For example, RGB images provide rich color and detailed texture information but are sensitive to changes in illumination. In contrast, IR images are less affected by illumination variations and can capture consistent information under different lighting conditions. Depth images, on the other hand, offer structural insights by providing depth information, though they often lack textural details.
To further explore the similar visual characteristics within individual modalities, we conduct a correlation analysis to investigate the feature similarity between live and spoof classes within each modality. Specifically, we use a pre-trained ResNet-34 model \cite{he2016deep} to extract latent features from the entire multi-modal dataset \textbf{CeFA} \cite{liu2021casia}, and then calculate the cosine similarity of the extracted features for the same modality in both the live and spoof classes.
As shown in Figure~\ref{fig:single}, features within the same modality of the live class demonstrate high similarity and exhibit smaller visual differences compared to those of the spoof class.

\begin{figure}  
    \centering
    \begin{tabular}{c }
        \begin{minipage}{0.9\columnwidth}
            \includegraphics[width=\linewidth]{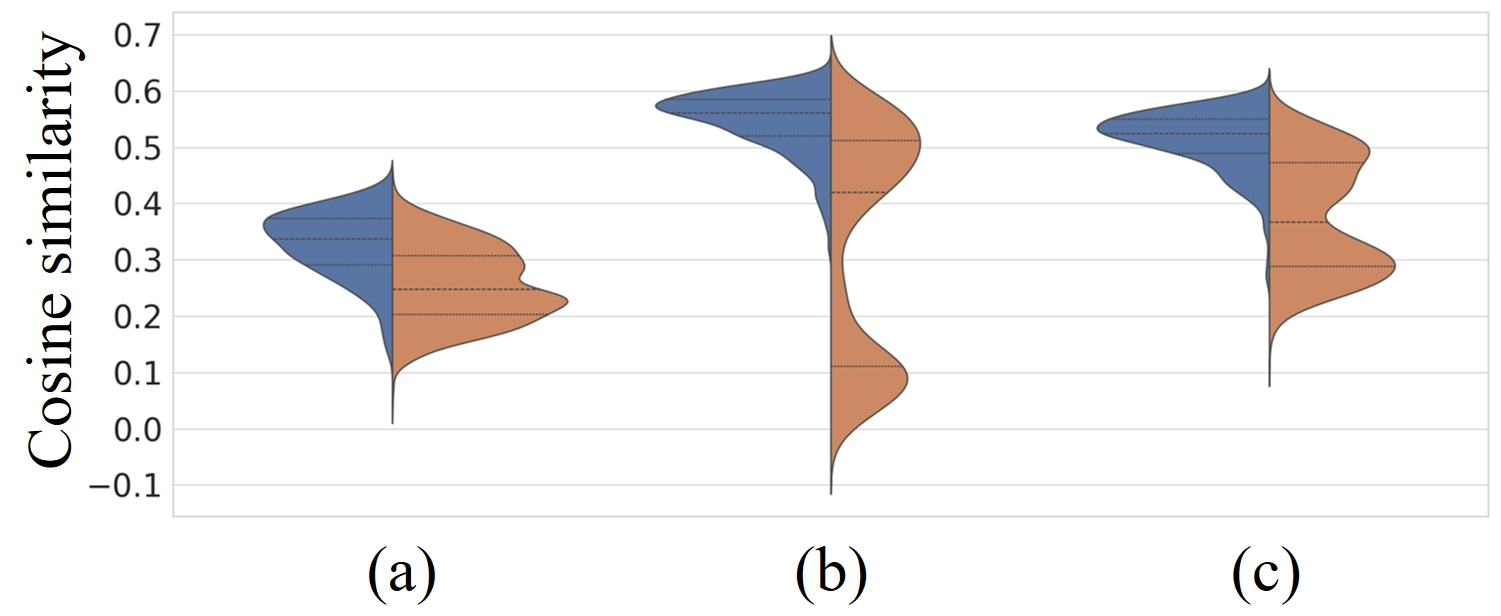}
        \end{minipage} 
    \end{tabular}
\caption{   
Distributions of the average Pearson Correlation for directional feature transitions across different modalities, including (a) RGB-IR, (b) RGB-D, and (c) IR-D transitions, for the live class (blue) and between live and spoof class (orange) in the \textbf{CeFA} dataset \cite{liu2021casia}.
Different feature transitions in the live class still exhibit relatively higher correlation compared to those between live and spoof classes. }  
\label{fig:transition} 
\end{figure}

\subsection{  Consistent Live Feature Transitions and Inconsistent Live and Spoof  Feature Transition across Modalities }  
\label{sec:feature transitions}
   
As mentioned above, since features of the live class within the same modality exhibit high similarity, we expect that the feature transitions across different modalities for samples from the live class should also be consistent with each other. For example, the differences in feature transitions from live RGB images to live IR images should be minimal.
To verify this, in Figure~\ref{fig:transition}, we further conduct a correlation analysis on the feature transitions between different modalities in the latent feature space for both live and spoof classes. Using the same feature extraction method described in Section \ref{sec: similar_live  feature}, we calculate the average Pearson Correlation    $\mathbf{r}_{\text{RGB} \rightarrow \text{IR}}$, $\mathbf{r}_{\text{RGB} \rightarrow \text{D}}$, and $\mathbf{r}_{\text{D} \rightarrow \text{IR}}$ of the feature transitions between different modalities within the live class for the $i$-th sample by, 
\begin{eqnarray}
    \mathbf{r}_{\text{m}_1 \rightarrow \text{m}_2} &=& \frac{1}{\left| S^l \right|} \sum_{\substack{\forall  \mathbf{f}^{\prime}_{\text{m}_1}, \mathbf{f}^{\prime}_{\text{m}_2} \in S^l \setminus \{ \mathbf{f}_{\text{m}_1}, \mathbf{f}_{\text{m}_2} \}, \\ \text{m}_1 \in \{ \text{RGB, IR} \}, \text{m}_2 \in \{ \text{IR, D} \}, \text{m}_1 \neq \text{m}_2}} \text{Pearson} ((\mathbf{f}_{\text{m}_2} - \mathbf{f}_{\text{m}_1}),(\mathbf{f}^{\prime}_{\text{m}_2} - \mathbf{f}^{\prime}_{\text{m}_1})) \nonumber \\
    &=& \frac{1}{\left| S^l \right|} \sum_{\substack{\forall 
 \mathbf{f}^{\prime}_{\text{m}_1}, \mathbf{f}^{\prime}_{\text{m}_2} \in S^l \setminus \{ \mathbf{f}_{\text{m}_1}, \mathbf{f}_{\text{m}_2} \}, \\ \text{m}_1 \in \{ \text{RGB, IR} \},  \text{m}_2 \in \{ \text{IR, D} \}, \text{m}_1 \neq \text{m}_2}} \text{Pearson} ((\mathbf{f}_{{\text{m}_1}\rightarrow{\text{m}_2}}),(\mathbf{f}^{\prime}_{{\text{m}_1}\rightarrow{\text{m}_2}}))
\end{eqnarray}

\noindent  
where $\mathbf{r}_{\text{m}_1 \rightarrow \text{m}_2}$ denotes the average Pearson Correlation of the feature transitions between modalities $\text{m}_1$ and  $\text{m}_2$ from the same class, $\text{m}_1 \in \{ \text{RGB, IR} \} $ and $ \text{m}_2 \in \{ \text{IR, D} \}$ represent specific modalities, $S^l = \{ \textbf{f}_{\text{RGB}}^i,  \textbf{f}_{\text{IR}}^i,  \textbf{f}_{\text{D}}^i\}_{i=1}^{ | S^l|}$ denotes the feature set of the all live training samples, $| S^l|$ denotes the size of $S^l$,
$\mathbf{f}_{{\text{m}_1}\rightarrow{\text{m}_2}}= \mathbf{f}_{\text{m}_2} - \mathbf{f}_{\text{m}_1}$ and $\mathbf{f}^{\prime}_{{\text{m}_1}\rightarrow{\text{m}_2}}= \mathbf{f}^{\prime}_{\text{m}_2} - \mathbf{f}^{\prime}_{\text{m}_1}$ represent the directional feature transitions vectors, 
$\mathbf{f}^{\prime}_{\text{m}_1}$  and $\mathbf{f}^{\prime}_{\text{m}_2}$ denote the live features of modalities $\text{m}_1$ and  $\text{m}_2$, 
$\mathbf{f}_{\text{m}_1}$ and $\mathbf{f}_{\text{m}_2}$ denote the latent features of the $i$-th sample, 
and $\text{Pearson}(\mathbf{a} , \mathbf{b})=  \frac{Cov(\mathbf{a}, \mathbf{b} )}{\sqrt{Cov(  \mathbf{a} ,  \mathbf{a})}\sqrt{Cov( \mathbf{b}, \mathbf{b} )}}$ measures the Pearson Correlation, where $Cov(\mathbf{a} , \mathbf{b} )$ denotes  the covariance between $\mathbf{a}$ and $\mathbf{b}$.  
As shown in Figure~\ref{fig:transition}, different feature transitions in the live class still exhibit relatively higher correlation compared to those between live and spoof classes. These findings indicate that consistent live feature transitions and inconsistent live and spoof feature transitions across modalities are indeed distinguishing characteristics of multi-modal FAS. 

\begin{figure*}
  \includegraphics[width= \linewidth]{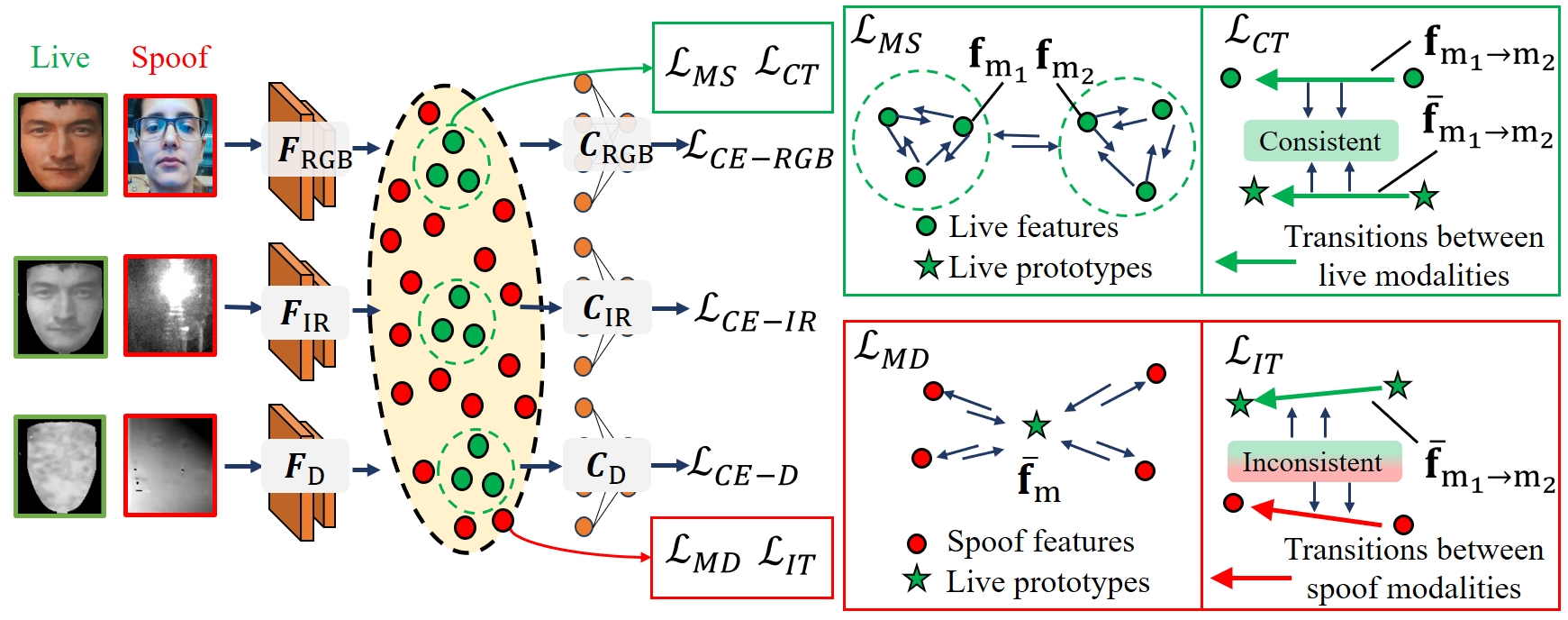}
  \caption{  
  The proposed CTNet aims to learn the consistent feature transitions from live samples and the consistent feature transitions between live and spoof samples  to facilitate the detection of out-of-distribution (OOD) occurrences during inference.
  } 
  \label{fig:framework}
\end{figure*}

\section{Proposed Method} 
In this paper, we propose a novel multi-modal face anti-spoofing (FAS) model, Cross-modal Transition-guided Network (CTNet), by leveraging the above-mentioned distinct characteristics of live and spoof classes across different modalities. Figure~\ref{fig:framework} shows the main idea of the proposed CTNet, which is developed to consistent cross-modal feature transitions among live samples, while learning the inconsistent cross-modal feature transitions between live and spoof samples to construct a generalized feature space.
To address the challenge of missing modalities during testing, we further propose an effective complementary feature learning approach. Through this method, the model is able to derive IR-like and depth-like features directly from RGB images to compensate the absence of IR or depth modalities. 
In Section~\ref{sec:CT-guided feature learning}, we propose a novel cross-modal transition-guided feature learning to learn feature transitions across different modalities for both live and spoof classes.
Next, in Section~\ref{sec:Complementary},  we propose an  effective complementary feature learning to tackle the challenge  of missing  modalities. 
Furthermore, in Section~\ref{sec:Total loss}, we summarize the total loss in this paper. 
Finally, in Section~\ref{sec:inference}, we introduce the detection score used in inference stage.

\subsection{ Cross-Modal Transition-Guided Feature Learning} 
\label{sec:CT-guided feature learning}
 
In Section~\ref{sec:Two Useful Properties}, we first introduce two useful properties for multi-modal face anti-spoofing.
Next, in Section~\ref{sec:Modality-GeneralizedFeature Learning}, we propose learning consistent feature transitions across different modalities for live class.
Finally, in Section~\ref{sec:MA Feature Learning}, we propose learning inconsistent feature transitions across different modalities between live and spoof classes.

\subsubsection{ The Two Useful Properties for Multi-Modal Face Anti-Spoofing}
\label{sec:Two Useful Properties}
 
Given a  multi-modal FAS training dataset $D = \{ \mathbf{x}_\text{RGB}, \mathbf{x}_\text{IR}, \mathbf{x}_\text{D}, y \}$, where $\mathbf{x}_\text{RGB}$, $\mathbf{x}_\text{IR}$, and $\mathbf{x}_\text{D}$ denote the RGB, IR, and depth modalities, respectively,  and $y$ denotes the binary live/spoof labels.
We denote the proposed Cross-modal Transition-guided Network (CTNet) by $T$. As the model $T$ is constructed with three feature extractors (i.e., $F_\text{RGB}$, $F_\text{IR}$, and $F_\text{D}$) and three classifiers (i.e., $C_\text{RGB}$, $C_\text{IR}$, and $C_\text{D}$), as shown in Figure~\ref{fig:framework}, we adopt exponential moving average to update the live modality prototypes from live training samples within current batch by, 
\begin{eqnarray} 
\bar{\mathbf{f}}_\text{m}  &=&  (1-\gamma)\,  \bar{\mathbf{f}}_\text{m}^\text{pre}  + \gamma \, \bar{\mathbf{f}}_\text{m}^\text{cur} 
\label{eq:prototypes_update}
\end{eqnarray}
where
\begin{eqnarray}
\bar{\mathbf{f}}_\text{m}^\text{cur} &=& \frac{1}{\left| B^{l} \right|}\sum_{\substack{\forall  \mathbf{f}_\text{m} \in B^l, \\ \text{m} \in \{\text{RGB}, \text{IR}, \text{D}\}}}\mathbf{f}_\text{m}
\label{eq:f_bar}  
\end{eqnarray}  

\noindent 
denotes the live modality prototype $\bar{\mathbf{f}}_\text{m}^\text{cur}$ calculated from current batch-wise live feature set $B^{l}$ for the modality $\text{m} \in \{\text{RGB}, \text{IR}, \text{D}\}$,  
 $\mathbf{f}_\text{m}$ denotes the modality-specific features extracted from the specific modality $\mathbf{x}_\text{m}$ (e.g.,  $\mathbf{f}_\text{RGB} = F_\text{RGB} (\mathbf{x}_\text{RGB})$), 
$\bar{\mathbf{f}}_\text{m}^\text{pre}$ and $\bar{\mathbf{f}}_\text{m} $ are the live modality prototypes before and after the update, respectively,  and $\gamma$ is a small weighted factor.

First, recall that features within the same modality of the live class exhibit high similarity and display smaller visual differences compared to those of the spoof class, as discussed in Section~\ref{sec: similar_live  feature}. 
Thus, it is reasonable to expect that the features of the live class cluster more closely around the live prototypes compared to those of the spoof class. 
We can leverage this property to distinguish live and spoof classes by, 
\begin{eqnarray}
 y &=& \left\{ \begin{array}{ll}
    1, & \text{if} \sum\limits_{\forall \text{m} \in \{ \text{RGB, IR, D} \}} \text{cos}(\bar{\mathbf{f}}_\text{m}, \mathbf{f}_\text{m}) \geq \alpha ; \\
    0, & otherwise.\\
\end{array} \right.
\label{eq:feature_similarity}
\end{eqnarray}

\noindent 
where $\text{cos}(\cdot)$ measures the cosine similarity, $\alpha$ is a small-value threshold, and $y = 1$ and $y = 0$ denote the live and spoof classes, respectively. 
If a sample belongs to the live class, its features ( $\mathbf{f}_\text{RGB}$, $\mathbf{f}_\text{IR}$, and $\mathbf{f}_\text{D}$) are expected to be close to the live prototypes, i.e., $\text{cos}( \bar{\mathbf{f}}_\text{RGB}, \mathbf{f}_\text{RGB})  + \text{cos}( \bar{\mathbf{f}}_\text{IR} , \mathbf{f}_\text{IR} )  + \text{cos}( \bar{\mathbf{f}}_\text{D} , \mathbf{f}_\text{D} )  \geq \alpha$.
By contrast, if a sample belongs to the spoof class, its features are expected to be distant from the live prototypes, i.e., $ \text{cos}( \bar{\mathbf{f}}_\text{RGB} , \mathbf{f}_\text{RGB})  + \text{cos}( \bar{\mathbf{f}}_\text{IR} , \mathbf{f}_\text{IR} )  + \text{cos}( \bar{\mathbf{f}}_\text{D} , \mathbf{f}_\text{D} )  < \alpha$.  

Next, in Section~\ref{sec:feature transitions},  the findings suggest that the consistent cross-modal feature transitions of live class and the inconsistent cross-modal feature transitions between live and spoof classes are key distinguishing characteristics of multi-modal FAS.
Thus, it is reasonable to measure the similarity of feature transitions between the live prototypes and the current sample to differentiate between live and spoof classes by, 
\begin{eqnarray}
y = \begin{cases} 
1, & \text{if} \sum\limits_{\substack{ \forall \text{m}_1 \in \{ \text{RGB, IR} \}, \\ \forall \text{m}_2 \in \{ \text{IR, D} \},  \text{m}_1 \neq \text{m}_2}} \text{Pearson} ((\bar{\mathbf{f}}_{{\text{m}_1}\rightarrow{\text{m}_2}}),(\mathbf{f}_{{\text{m}_1}\rightarrow{\text{m}_2}}))\geq \beta; \\
0, & otherwise.
\end{cases}
\label{eq:transition_similarity}
\end{eqnarray}

\noindent 
where $\bar{\mathbf{f}}_{{\text{m}_1}\rightarrow{\text{m}_2}}$ and $\mathbf{f}_{{\text{m}_1}\rightarrow{\text{m}_2}}$ represent  the feature transitions of the live prototypes and the current sample, respectively,  and  $\beta $ is a margin to distinguish live and spoof class. 
If a sample belongs to the live class, its feature transitions are expected to be close to the feature transitions of the live prototypes.
By contrary, if a sample belongs to the spoof class, its feature transitions are expected to be distant from the feature transitions of the live prototypes.
  
While the two properties in Equations~\ref{eq:feature_similarity} and \ref{eq:transition_similarity} are reasonable, they remain unexplored in the multi-modal FAS task.  
Therefore, we propose leveraging these two useful properties to develop a cross-modal transition-guided formulation of feature learning to  facilitate multi-modal FAS.  

\subsubsection{ Consistent Transition Feature Learning for Live Class} 
\label{sec:Modality-GeneralizedFeature Learning} 
In this Section, we propose learning consistent feature transitions  solely from samples of the live class to construct  a latent space that encapsulates the two aforementioned useful properties.
First, we propose to learn the modality-specific features $\mathbf{f}_{\text{m}_1}$ for the specific modality $\text{m}_1$.
In particular, we define the modality-specific loss $\mathcal{L}_{MS}$ to pull features of the same modality closer together  while pushing them away from features of different modalities from live training samples by, 
\begin{eqnarray} 
\mathcal{L}_{MS} =
\sum_{\substack{\forall \mathbf{f}_{\text{m}_1}, \mathbf{f}^{\prime}_{\text{m}_1} \in B^l,  \mathbf{f}_{\text{m}_1} \neq \mathbf{f}_{\text{m}_1}^{\prime}, \\\forall  \text{m}_1 \in \{ \text{RGB, IR,D} \} }}   
(  & - log \frac{\text{exp}(\text{cos}(\textbf{f}_{\text{m}_1},\textbf{f}_{\text{m}_1}^{\prime}))} { \sum\limits_{\substack{
\forall \mathbf{f}^{\prime}_{\text{m}_2} \in B^l,\\
\forall \text{m}_2 \in \{\text{RGB}, \text{IR}, \text{D}\}, \text{m}_2 \neq \text{m}_1}}  \text{exp}(\text{cos}(\textbf{f}_{\text{m}_1},\textbf{f}_{\text{m}_2}^{\prime})) }  )
\label{eq:ms_loss}
\end{eqnarray} 

\noindent 
where $B^l = \{ \textbf{f}_{\text{RGB}}^i,  \textbf{f}_{\text{IR}}^i,  \textbf{f}_{\text{D}}^i\}_{i=1}^{ | B^l|}$ denotes the live modality feature set of current batch, $ | B^l |$ is the size of $B^l$, and 
the denominator only includes the negative pairs to remove the negative-positive-coupling effect \cite{yeh2022decoupled}.
Next, we adopt Equation~\ref{eq:prototypes_update}  to update the live modality prototypes. 
After obtaining the updated live modality prototypes, our next goal is to learn a generalized feature space by learning the consistent feature transitions from live class. 
We define the consistent transition loss $\mathcal{L}_{CT}$ as follows:    
\begin{eqnarray}
\mathcal{L}_{CT} &=&   \sum_{\substack{\forall \mathbf{f}_{\text{m}_1}, \mathbf{f}_{\text{m}_2} \in B^l,    \\
 \forall \text{m}_1 \in \{ \text{RGB, IR} \}, \\ \forall \text{m}_2 \in \{ \text{IR, D} \} , \text{m}_1 \neq \text{m}_2
}} 
 ( 1 - \text{Pearson}(\mathbf{f}_{{\text{m}_1}\rightarrow{\text{m}_2}}, \bar{\mathbf{f}}_{{\text{m}_1}\rightarrow{\text{m}_2}})).
\label{eq:MT}
\end{eqnarray}

\subsubsection{ Inconsistent Transition Feature Learning between Live and Spoof Class} 
\label{sec:MA Feature Learning}
  
Thus far, we have established a consistent transition feature learning formulation to derive a generalized feature space exclusively from the training samples of the live class.
However, the feature learning approach described in Section \ref{sec:Modality-GeneralizedFeature Learning} does not delve into the distinctive characteristics of the live and spoof classes within the latent feature space.
Building on the findings from Sections~\ref{sec:feature transitions} and \ref{sec:Two Useful Properties}, we are now prepared to incorporate spoof training samples to learn the inconsistent feature transitions between the live and spoof classes  for constructing the discriminative feature space. 
In particular, with the updated live prototypes from Equation~\ref{eq:prototypes_update}, our first goal is to learn  discriminative spoof features by pushing spoof features distinctly away from the live prototypes. 
Therefore, we define the modality-discriminative loss $\mathcal{L}_{MD}$  to learn the discriminative spoof features by,  
\begin{eqnarray}
\mathcal{L}_{MD} = 
\sum\limits_{\substack{\forall \mathbf{f}_\text{m} \in B^s, \\ \forall \text{m} \in \{ \text{RGB, IR, D} \}}} \text{cos}(\bar{\mathbf{f}}_\text{m}, \mathbf{f}_\text{m}),
\label{eq:MD}
\end{eqnarray}

\noindent 
where $B^s = \{ \textbf{f}_{\text{RGB}}^i,  \textbf{f}_{\text{IR}}^i,  \textbf{f}_{\text{D}}^i\}_{i=1}^{ | B^s|}$ denotes the spoof modality feature set of current batch and $ | B^s |$ is the size of $B^l$. 
 
Next, to facilitate detecting Out-of-Distribution (OOD) attacks in inference, our second goal is to learn the inconsistent feature transitions between the live and spoof classes.
Thus, we define the inconsistent transition loss $\mathcal{L}_{IT}$ to 
push the cross-modal feature transitions of the spoof class away from the cross-modal feature transitions of the live prototypes by,  
\begin{eqnarray}
    \mathcal{L}_{IT} &=& \sum_{\substack{ \forall \mathbf{f}_{\text{m}_1}, \mathbf{f}_{\text{m}_2} \in B^{s},  \\ \forall \text{m}_1 \in \{ \text{RGB, IR} \}, \\ \forall \text{m}_2 \in \{ \text{IR, D} \}, \text{m}_1 \neq \text{m}_2}} \text{Pearson} ((\mathbf{\bar{f}}_{{\text{m}_1}\rightarrow{\text{m}_2}}),(\mathbf{f}_{{\text{m}_1}\rightarrow{\text{m}_2}})).
\label{eq:TD}
\end{eqnarray}

In addition, we also include the cross entropy losses  $\mathcal{L}_{CE-RGB}$, $\mathcal{L}_{CE-IR}$, and $\mathcal{L}_{CE-D}$ to learn liveness information from each modality.

\subsection{Complementary Feature Learning}
\label{sec:Complementary}  
To tackle the issue of missing modalities (e.g., IR or depth modalities) during inference stage, we propose an effective complementary feature learning to extract  the IR-like and depth-like features as    complementary representations when IR or depth modalities are unavailable.
In particular, we additionally include the two auxiliary feature extractors $\hat{F}_\text{IR}$, $\hat{F}_\text{D}$, and define the complementary feature loss $\mathcal{L}_{CF}$  to learn the IR-like features $\hat{\mathbf{f}}_\text{IR}$ and the depth-like features $\hat{\mathbf{f}}_\text{D}$ from the all RGB images by,  
\begin{eqnarray}
    \mathcal{L}_{CF} &=& \sum\limits_{\substack{\forall \mathbf{f}_\text{m} \in B^{ls}, \\ \forall \text{m} \in \{ \text{IR, D} \}}} (1 - \text{cos}(\mathbf{f}_\text{m}, \hat{\mathbf{f}}_\text{m}))
\label{eq:cf}
\end{eqnarray}

\noindent  
where $ B^{ls} =  B^{l} \cup  B^{s}$ present the live and spoof feature set of current batch, 
$\hat{\mathbf{f}}_\text{IR} = \hat{F}_\text{IR} (\mathbf{x}_\text{RGB})$ and $\hat{\mathbf{f}}_\text{D} = \hat{F}_\text{D} (\mathbf{x}_\text{RGB})$ are the IR-like and depth-like features, respectively.
When encountering missing modalities during the inference stage, we propose to use $\hat{\mathbf{f}}_\text{IR}$ and $\hat{\mathbf{f}}_\text{D}$ as replacements for $\mathbf{f}_\text{IR}$ and $\mathbf{f}_\text{D}$, respectively.  

\subsection{Total loss}
\label{sec:Total loss}
  
Finally, we include the  modality-specific loss $\mathcal{L}_{MS}$, the consistent transition loss $\mathcal{L}_{CT}$, the modality-discriminative loss $\mathcal{L}_{MD}$, the inconsistent transition loss $\mathcal{L}_{AT}$, the complementary feature loss $\mathcal{L}_{CF}$, and  the cross entropy losses  $\mathcal{L}_{CE-RGB}$, $\mathcal{L}_{CE-IR}$, and $\mathcal{L}_{CE-D}$  to define the total loss $\mathcal{L}_T$ by, 
\begin{eqnarray}
\mathcal{L}_T &=& \mathcal{L}_{MS} + \mathcal{L}_{CT} + \lambda_1 \mathcal{L}_{MD} + \lambda_1 \mathcal{L}_{IT} + \mathcal{L}_{CF} +
\nonumber \\
&& \lambda_2 \mathcal{L}_{CE-RGB} + \lambda_2  \mathcal{L}_{CE-IR} + \lambda_2  \mathcal{L}_{CE-D},
\label{eq:total_loss}  
\end{eqnarray}
\noindent  
where $\lambda_1=0.005$ and $\lambda_2=0.5$ are the weighted factors.
Note that, whether $\mathcal{L}_{CF}$ is included in $\mathcal{L}_{T}$ depends on the testing scenarios. 
For example, in the fixed-modal scenario (where testing data includes entire RGB, IR, and depth images), $\mathcal{L}_{CF}$ would not be included in $\mathcal{L}_T$.
By contrary, in the missing-modal scenario (where testing data may lack  IR or depth images), $\mathcal{L}_{CF}$ will be included in $\mathcal{L}_T$.
In addition, the two auxiliary feature extractors $\hat{F}_\text{IR}$, $\hat{F}_\text{D}$ will be included to assist CTNet  under the missing-modal scenario. 

\subsection{Live/Spoof Classification }
\label{sec:inference}
 
In inference, we adopt the trained CTNet to the test images  $\mathbf{x}_{\text{RGB}}$, $\mathbf{x}_{\text{IR}}$, and $\mathbf{x}_{\text{D}}$  and calculate its OOD score  $SC_{ood}$ by,  
\begin{eqnarray} 
SC_{ood} =  (1-\lambda_3) SC_t + \lambda_3 SC_d, 
\label{eq:s_all}
\end{eqnarray}

\noindent
which incorporates both the distance score $SC_d$ and the transition score $SC_t$ defined by, 
\begin{eqnarray} 
SC_d = \sum_{\forall \text{m} \in \{\text{RGB}, \text{IR}, \text{D}\}} (1-\text{cos}(\bar{\mathbf{f}}_\text{m} , \mathbf{f}_\text{m})) ,
\label{eq:score_d}
\end{eqnarray}
and
 \begin{eqnarray} 
SC_t = \sum_{\substack{ \forall \text{m}_1 \in \{ \text{RGB, IR} \}, \\ \forall \text{m}_2 \in \{ \text{IR, D} \}, \text{m}_1 \neq \text{m}_2}} (1-\text{Pearson} ((\mathbf{\bar{f}}_{{\text{m}_1}\rightarrow{\text{m}_2}}),(\mathbf{f}_{{\text{m}_1}\rightarrow{\text{m}_2}})))
\label{eq:socre_s}
\end{eqnarray}

\noindent
\textcolor{black}{
where $\lambda_3 \in [0,1]$ is the weighted factor. Note that, we follow previous methods \cite{wang2022domain, yu2020searching} to adopt Youden Index Calculation \cite{youden1950index} for obtaining the threshold of binary classification.
}

\section{Experiment}

\subsection{Experimental setting}
\subsubsection{Datasets}
 
We conduct extensive experiments on the following three multi-modal face anti-spoofing databases: \textbf{WMCA} \cite{george2019biometric} (denoted by \textbf{W}), \textbf{CASIA-SURF} \cite{zhang2020casia} (denoted by \textbf{S}), and \textbf{CASIA-SURF CeFA} \cite{liu2021casia} (denoted by \textbf{C}).
In particular, \textbf{WMCA} \cite{george2019biometric}  consists of seven types of presentation attacks, including both 2D  and 3D attack types. 
Next, \textbf{CASIA-SURF} \cite{zhang2020casia} comprises 21,000 videos featuring 1,000 subjects, including print attacks, some of which involve partially cut-out facial features.
Furthermore, \textbf{CASIA-SURF CeFA} \cite{liu2021casia} includes 1,607 subjects from three different ethnicities and provides both 2D and 3D presentation attacks. It also includes protocols for testing unseen presentation attacks and cross-ethnicity subjects.

\subsubsection{Evaluation metrics}
  
To ensure a fair comparison with previous multi-modal face anti-spoofing methods, we  employ metrics such as the Attack Presentation Classification Error Rate (APCER) (\%) $\downarrow$ \cite{standard2016information}, Bona Fide Presentation Classification Error rate (BPCER) (\%) $\downarrow$ \cite{standard2016information}, and Average Classification Error Rate (ACER) (\%) $\downarrow$ \cite{standard2016information} by, 
\begin{eqnarray} 
 \text{APCER} = \frac{\text{FP}}{\text{FP}+\text{TN}},
\end{eqnarray}
\begin{eqnarray} 
 \text{BPCER} = \frac{\text{FN}}{\text{FN}+\text{TP}},
\end{eqnarray}
\begin{eqnarray} 
 \text{ACER} = \frac{\text{APCER} + \text{BPCER}}{2},
\end{eqnarray}

\noindent 
where TP, TN, FP and FN present True Positive, True Negative,
False Positive and False Negative, respectively. 
In particular, as noted in \cite{huang2024survey}, APCER measures the error rate of incorrectly classifying spoof attempts as genuine, while BPCER evaluates the error rate of rejecting legitimate users.  
By combining APCER and BPCER, ACER provides a valuable overall performance indicator. 
In addition, we also adopt  Area Under Curve (AUC) (\%) $\uparrow$ to evaluate the model's ability to distinguish genuine users from spoof attacks across different thresholds. 

\subsubsection{Implementation details}
\label{sec: Implementation}
 
We use the pre-trained ResNet-34 as the backbones for the proposed CTNet.
To train CTNet, we set a constant learning rate of 5e-4 with AdamW optimizer up to 50 epochs.
In \cite{yu2023flexible}, the authors introduced the fixed-modal scenario (where testing data includes the complete set of RGB, IR, and depth images) and the missing-modal scenario (where testing data may be missing IR or depth images).
Given the flexibility of the proposed CTNet to support both fixed- and  missing-modal  scenarios, we employ different losses to train CTNet accordingly under each setting. 
In particularly, we adopt the same operation mentioned in Section~\ref{sec:Total loss} to train CTNet for both the fixed-modal and missing-modal scenarios.

\subsection{Ablation study}

\begin{table}[t]
\caption{  
Ablation study on  the cross-domain protocol 1 of the missing-modal scenario \cite{yu2023flexible}, using different loss combinations.
} 
 \centering
 \label{tab:ablation_different_losses}
\begin{tabular}{cccccc|cc}
\hline
\multicolumn{6}{c|}{Loss Term}                                                                                                             & \multicolumn{2}{c}{Protocol 1}                    \\ \hline
\multicolumn{1}{c|}{\makecell{$ \mathcal{L}_{CE}$}}  & \multicolumn{1}{c|}{$\mathcal{L}_{CF}$}  & \multicolumn{1}{c|}{$\mathcal{L}_{MS}$}  & \multicolumn{1}{c|}{$\mathcal{L}_{MD}$}  & \multicolumn{1}{c|}{$\mathcal{L}_{CT}$}  & $\mathcal{L}_{IT}$  & \multicolumn{1}{c|}{ACER $\downarrow$} & AUC  $\uparrow$                  \\ \hline
\multicolumn{1}{c|}{\checkmark } & \multicolumn{1}{c|}{}    & \multicolumn{1}{c|}{}    & \multicolumn{1}{c|}{}    & \multicolumn{1}{c|}{}    &     & \multicolumn{1}{c|}{35.12}     &  70.95                    \\ \hline
\multicolumn{1}{c|}{\checkmark } & \multicolumn{1}{c|}{\checkmark } & \multicolumn{1}{c|}{}    & \multicolumn{1}{c|}{}    & \multicolumn{1}{c|}{}    &     & \multicolumn{1}{c|}{32.11 }     &     73.09                 \\ \hline
\multicolumn{1}{c|}{\checkmark } & \multicolumn{1}{c|}{\checkmark } & \multicolumn{1}{c|}{\checkmark } & \multicolumn{1}{c|}{}    & \multicolumn{1}{c|}{}    &     & \multicolumn{1}{c|}{30.44}     & 74.04                     \\ \hline
\multicolumn{1}{c|}{\checkmark } & \multicolumn{1}{c|}{\checkmark } & \multicolumn{1}{c|}{\checkmark } & \multicolumn{1}{c|}{ } & \multicolumn{1}{c|}{\checkmark}    &     & \multicolumn{1}{c|}{27.70}     &     78.46                \\ \hline
\multicolumn{1}{c|}{\checkmark } & \multicolumn{1}{c|}{\checkmark } & \multicolumn{1}{c|}{\checkmark } & \multicolumn{1}{c|}{\checkmark } & \multicolumn{1}{c|}{\checkmark } &     & \multicolumn{1}{c|}{25.04}     &      80.00                \\ \hline
\multicolumn{1}{c|}{\checkmark } & \multicolumn{1}{c|}{\checkmark } & \multicolumn{1}{c|}{\checkmark } & \multicolumn{1}{c|}{\checkmark } & \multicolumn{1}{c|}{\checkmark } & \checkmark  & \multicolumn{1}{c|}{ \textbf{22.56} }   & \textbf{81.78} \\ \hline
\end{tabular}
\end{table}

\subsubsection{ Comparison between different loss terms}
 
In Table~\ref{tab:ablation_different_losses}, we compare using different loss terms to train the proposed CTNet on  the cross-domain protocol 1 of the missing-modal scenario \cite{yu2023flexible}, and using the detection scores $SC_{ood}$ in the inference stage. 
 
First, we adopt a simple combination of cross-entropy losses, $\mathcal{L}_{CE} = \mathcal{L}_{CE-RGB} + \mathcal{L}_{CE-IR} + \mathcal{L}_{CE-D}$, to train the FAS model as the baseline. We follow previous method to adopt zero-padding for missing modalities as input during the inference stage in the baseline method.
Due to the missing modalities, we observe poor performance in the baseline method. 
 
Next, we include $\mathcal{L}_{CF}$  to train the FAS model. 
Because $\mathcal{L}_{CF}$ constrain the FAS model to extract the complementary IR-like and depth-like features from RGB modality, these complementary features are able to effectively address the challenge posed by the missing modalities to enhance the performance of the FAS model. 
 
Furthermore, we include $\mathcal{L}_{MS}$ as well as  $\mathcal{L}_{CT}$ to train the FAS model. 
By comparing the cases of  $\mathcal{L}_{CE}+\mathcal{L}_{CF}$ and $\mathcal{L}_{CE}+\mathcal{L}_{CF}+\mathcal{L}_{MS}$,  we observe a slight improvement in performance, attributed to the learning of modality-specific features from each modality.
By comparing the cases of $\mathcal{L}_{CE}+\mathcal{L}_{CF}+\mathcal{L}_{MS}$ and $\mathcal{L}_{CE}+\mathcal{L}_{CF}+\mathcal{L}_{MS}+\mathcal{L}_{CT}$, we observe that learning consistent feature transitions between live samples enhances the generalization ability and further improves performance. 
 
Moreover, we include $\mathcal{L}_{MD}$ to learn discriminative spoof features that are distinct from the features of different modalities in live training samples. By incorporating this, the model effectively separates spoof features from live ones, resulting in enhanced performance in detecting spoof attacks. 
 
Finally, by incorporating $\mathcal{L}_{IT}$, we achieve the best performance. The proposed CTNet indeed benefits from learning inconsistent feature transitions between live and spoof samples, which significantly enhances its ability to detect out-of-distribution (OOD) attacks.

\subsubsection{ Comparison between different $\lambda_3$ in OOD score $SC_{ood}$
}

\begin{figure}
    \centering
    \includegraphics[width=0.9\columnwidth]{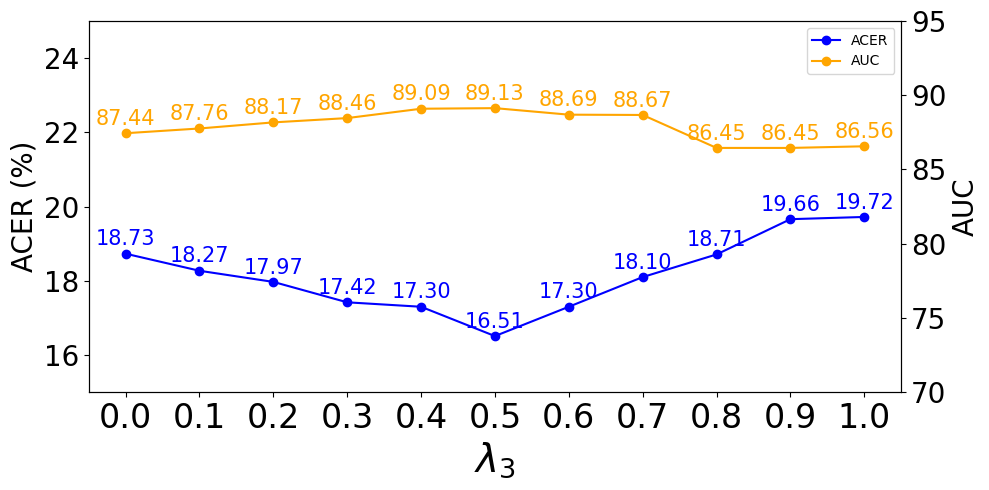}
    \caption{  Ablation study on  the cross-domain protocol 4 of the missing-modal scenario \cite{yu2023flexible}, using different weighting of $\lambda_3$ in Equation \eqref{eq:s_all}. }
    \label{fig:lambda3}
\end{figure}
 
In Figure~\ref{fig:lambda3},  we compare the performance of using different weighted factors $\lambda_3$ to fuse the two detection scores, $SC_t$ and $SC_d$, for calculating the OOD score $S_{ood}$ in live/spoof classification.
First, by comparing the performances of $\lambda_3=0$ (i.e., using $SC_t$) and $\lambda_3=1$ (i.e., using $SC_d$), we observe that only measuring transition consistency to distinguish live and spoof faces outperforms solely measuring the distances between the live prototypes and the modality-specific features.
Next, we observe that the performance steadily improves as $\lambda_3$ increases from 0 to 0.5, achieving the best ACER and AUC at $\lambda_3 = 0.5$. However, the performance does not show further improvement as $\lambda_3$ increases beyond 0.6.
By comparing the cases of $\lambda_3=0$, $\lambda_3=1$, and other values of $\lambda_3$, we see that adaptive fusion indeed yields superior performance compared to utilizing a single score (i.e., $\lambda_3=1$ for $S_d$  or  $\lambda_3=0$ for $S_t$ ).  
Therefore, from this ablation study, we empirically set $\lambda_3=0.5$ for  subsequent experiments.

\begin{table}
    \caption{  
    Ablation study on  the cross-domain protocol 4 of the missing-modal scenario \cite{yu2023flexible}, using different transition constraint. 
    } 
    \centering
    \label{tab:ablation_two-class_transition}
        \begin{tabular}{cccc|c|c}
            \toprule
            \multicolumn{4}{c}{Loss Term} & \multicolumn{2}{c}{Protocol 4 } \\
            \midrule
            \multirow{2}{*}{\makecell{$ \mathcal{L}_{CE-RGB} +\mathcal{L}_{CE-IR}$ \\  $+\mathcal{L}_{CE-D} +\mathcal{L}_{MS} +\mathcal{L}_{MD}$}} & \multicolumn{3}{|c|}{$\mathcal{L}_{CT} +\mathcal{L}_{AT}$} & \multirow{2}{*}{ACER $\downarrow$}   & \multirow{2}{*}{AUC  $\uparrow$} \\
            \cmidrule{2-4}
             & \multicolumn{1}{|c}{$\mathbf{f}_{{\text{RGB}}\rightarrow{\text{IR}}}$} & $\mathbf{f}_{{\text{RGB}}\rightarrow{\text{D}}}$ & $\mathbf{f}_{{\text{IR}}\rightarrow{\text{D}}}$& \\
             \midrule
             \checkmark & \checkmark  & & & 19.82 & 85.60 \\
             \checkmark & & \checkmark &  & 18.90 & 87.70 \\
             \checkmark & & & \checkmark  & 17.92 & 87.41 \\
             \checkmark & \checkmark  & \checkmark & \checkmark & \textbf{16.51 }& \textbf{90.31} \\
            \bottomrule
        \end{tabular}
\end{table}

\subsubsection{Comparison between different combinations of feature transitions }
 
In Table~\ref{tab:ablation_two-class_transition}, we use the different combinations of feature transitions within  the losses  $\mathcal{L}_{CT}$ and $\mathcal{L}_{IT}$ to investigate the impact of  feature transitions on the overall performance of the proposed CTNet. 
 
First, we  only use individual feature transition (e.g., $\mathbf{f}_{\text{RGB}\rightarrow\text{IR}}$,  $\mathbf{f}_{\text{RGB}\rightarrow\text{D}}$, or $\mathbf{f}_{\text{D}\rightarrow\text{IR}}$) to calculate the loss constant within $\mathcal{L}_{CT}$ and $\mathcal{L}_{IT}$, and then compute  the corresponding  the detection score $SC_t$  based solely on the individual feature transition for inclusion in the OOD score   $SC_{ood}$. 
As shown in Figure \ref{fig:transition}, the differences of the RGB-IR feature transition between live and spoof classes are not significant.
Consequently, relying solely on the feature transition $\mathbf{f}_{\text{RGB}\rightarrow\text{IR}}$ to impose constraints results in poor performance, as reflected in Table~\ref{tab:ablation_two-class_transition}. 
 
Next, by comparing the cases of $\mathbf{f}_{\text{RGB}\rightarrow\text{IR}}$,  $\mathbf{f}_{\text{RGB}\rightarrow\text{D}}$, and $\mathbf{f}_{\text{D}\rightarrow\text{IR}}$, we show that  the significant differences of the RGB-D and IR-D feature transitions (as illustrated in Figure \ref{fig:transition} (b) and (c)) indeed encourage the proposed CTNet to focus on learning discriminative liveness features, thereby improving performance compared to the RGB-IR feature transition. 
 
Finally, by combining all transitions, we achieve the best performance, as this approach  probes the consistent feature transitions within the live class and the inconsistent feature transitions between live and spoof classes, thereby enhancing the ability of the proposed CTNet to effectively counter unseen spoof attacks.  

\begin{figure} [t] 
    \begin{minipage}[b]{1.0\linewidth}
      \centering{ 
      \includegraphics[width=0.6\columnwidth]{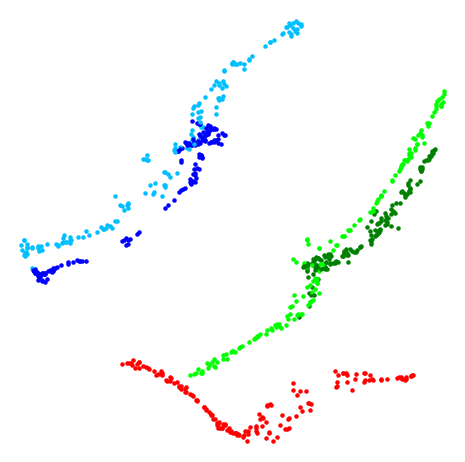}
      }
    \end{minipage} 
\caption{    
$t$-SNE visualizations on  the cross-domain protocol 4 of the missing-modal scenario \cite{yu2023flexible}, using the trained CTNet.
DOTS: modality-specific features (red: RGB features $ \textbf{f}_{\text{RGB}}$, dark green: IR features $ \textbf{f}_{\text{IR}}$, light green: IR-like features $ \hat{\textbf{f}}_{\text{IR}}$, blue: depth features $ \textbf{f}_{\text{D}}$, and cyan: depth-like features $ \hat{\textbf{f}}_{\text{D}}$).  
} 
\label{fig:tsne}  
\end{figure} 
 
\subsubsection{$t$-SNE visualization}
 
In Figure~\ref{fig:tsne}, we adopt $t$-SNE to visualize the latent liveness features on  the cross-domain protocol 4 of the missing-modal scenario \cite{yu2023flexible}.  
First, we observe that CTNet effectively extracts modality-specific features from different modalities, as represented by distinct clusters in the feature space (e.g., red dots for RGB features $ \textbf{f}_{\text{RGB}}$, dark green dots for IR features $ \textbf{f}_{\text{IR}}$, and blue dots for depth features $ \textbf{f}_{\text{D}}$).
Next, by comparing the IR features $ \textbf{f}_{\text{IR}}$ (dark green dots) vs.  the IR-like features $ \hat{\textbf{f}}_{\text{IR}}$ (light green dots) and the depth features $ \textbf{f}_{\text{D}}$ (blue dots) vs. the depth-like features $ \hat{\textbf{f}}_{\text{D}}$ (cyan dots), we observe that the IR-like and depth-like features extracted from RGB images closely align with the respective IR and depth features in the latent feature space.
This observation demonstrates that the proposed complementary feature learning effectively simulates the features of the missing modalities to facilitate the multi-modal face anti-spoofing task.

\begin{figure}  
    \centering
    \begin{tabular}{c }
        \begin{minipage}{0.95\columnwidth}
            \includegraphics[width=\linewidth]{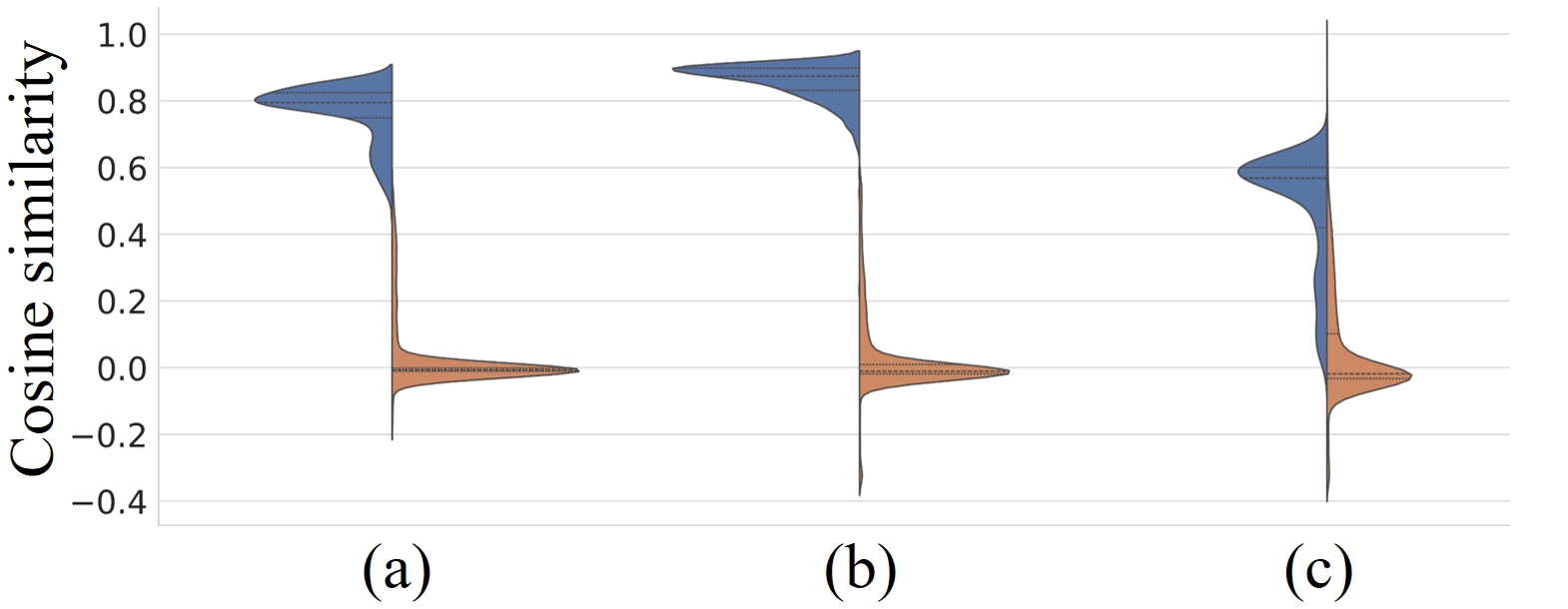}
        \end{minipage}  
    \end{tabular}
\caption{   
Distributions of the average Pearson Correlation for directional feature transitions across different modalities using the trained CTNet: (a) RGB-IR, (b) RGB-D, and (c) IR-D transitions, between live samples and live prototypes (blue) and between spoof samples and live prototypes (orange) in the \textbf{CeFA} dataset \cite{liu2021casia}.
}  
\label{fig:transition_trained} 
\end{figure}

\subsubsection{Correlation analysis of feature transitions}
 
In Figure~\ref{fig:transition_trained}, we use the trained CTNet to perform a correlation analysis on the feature transitions between different modalities within the latent feature space.
In particular, leveraging the consistent feature learning in Equation~\eqref{eq:MT} and the inconsistent feature learning in Equation~\eqref{eq:TD}, we measure the feature transitions between the features of the current sample and the live prototypes to evaluate their alignment and divergence on  the cross-domain protocol 4 of the missing-modal scenario \cite{yu2023flexible}.
From this analysis, the consistent feature transitions among live samples quantify the alignment and coherence of modality-specific features for live samples, while the inconsistent feature transitions between live and spoof samples validate the distinct deviation patterns characteristic of spoof samples.
Hence, by leveraging the characteristics of consistent and inconsistent feature transitions, the proposed CTNet provides a robust and insightful framework for effectively distinguishing live faces from spoof attacks, even under challenging conditions such as missing modalities and out-of-distribution attacks.

\begin{table*}[t]
\centering
\caption{ 
Intra-domain testing on \textbf{WMCA} \cite{george2019biometric} under the fixed-modal scenario. The evaluation metrics is ACER (\%) $\downarrow$.
} 
\label{tab:WMCA_intra}
\resizebox{\textwidth}{!}{
\begin{tabular}{c|c|cccc}
\toprule
\multirow{2}{*}{Method} & \multirow{2}{*}{Seen} & \multicolumn{4}{c}{Unseen}                                                         \\ \cline{3-6} 
&  & \multicolumn{1}{c}{Flexiblemask} & \multicolumn{1}{c}{Replay} & \multicolumn{1}{c}{Fakehead} & \multicolumn{1}{c}{Prints}  \\ \midrule

CMFL \cite{george2021cross}  (CVPR 21)                &   1.70              & 
                        \multicolumn{1}{c}{12.40}             & \multicolumn{1}{c}{1.00}       &
                        \multicolumn{1}{c}{2.50}         & \multicolumn{1}{c}{0.70}       \\

MA-ViT \cite{liu2023ma}  (IJCAI 22)              &   1.45                   & 
                        \multicolumn{1}{c}{9.76}             & 
                        \multicolumn{1}{c}{0.93}       & 
                        \multicolumn{1}{c}{0.55}         & 
                        \multicolumn{1}{c}{\textbf{0.00}}  
                        \\

Conv-MLP \cite{wang2022conv}  (TIFS 22)         &  -              & 
                        \multicolumn{1}{c}{12.60}             & \multicolumn{1}{c}{0.80}       &
                        \multicolumn{1}{c}{0.20}         & \multicolumn{1}{c}{0.10}       
                        \\
                        
FM-ViT \cite{liu2023fm}   (TIFS 23)                       &  1.0              & 
                        \multicolumn{1}{c}{3.56}       & 
                        \multicolumn{1}{c}{0.72}         & \multicolumn{1}{c}{\textbf{0.00}}       & 
                        \multicolumn{1}{c}{\textbf{0.00}}        
                        \\
LPST \cite{li2023learning}   (AAAI 23)             &  -              & 
                        \multicolumn{1}{c}{19.0}       & 
                        \multicolumn{1}{c}{0.5}         & \multicolumn{1}{c}{ 2.3}       & 
                        \multicolumn{1}{c}{ 0.7}        
                        \\
                        
ViT+AMA+M$^2$\text{A}$^2$\text{E} \cite{yu2024rethinking} (IJCV 24)             &  1.39              & 
                        \multicolumn{1}{c}{9.02}       & 
                        \multicolumn{1}{c}{\textbf{0.00}}         & \multicolumn{1}{c}{\textbf{0.00}}       & 
                        \multicolumn{1}{c}{\textbf{0.00}}       
                        \\

FM-CLIP \cite{liu2024fm} (ACMMM 24) &  1.05              & 
                        \multicolumn{1}{c}{3.35}       & 
                        \multicolumn{1}{c}{0.69}         & \multicolumn{1}{c}{\textbf{0.00}}       & 
                        \multicolumn{1}{c}{\textbf{0.00}}      
                        \\

CTNet (Ours)                     &    \textbf{0.98}                 & 
                        \multicolumn{1}{c}{\textbf{2.59}}             & 
                        \multicolumn{1}{c}{\textbf{0.00}}       & 
                        \multicolumn{1}{c}{\textbf{0.00}}         & 
                        \multicolumn{1}{c}{\textbf{0.00}}  
                        \\

\toprule

\multirow{2}{*}{Method} & \multirow{2}{*}{-} & \multicolumn{4}{c}{Unseen}                                                       \\ \cline{3-6} 
&  & \multicolumn{1}{c}{Glasses} & \multicolumn{1}{c}{Papermask} & \multicolumn{1}{c}{Rigidmask} & Mean \\ \midrule

CMFL \cite{george2021cross}  (CVPR 21) &       -              & 
                        \multicolumn{1}{c}{33.50}        & 
                        \multicolumn{1}{c}{1.80}          & \multicolumn{1}{c}{1.70}          &  
                        7.60
                        \\    
                        
MA-ViT \cite{liu2023ma}  (IJCAI 22)              &- &
                        \multicolumn{1}{c}{14.00}        & 
                        \multicolumn{1}{c}{\textbf{0.00}}          & 
                        \multicolumn{1}{c}{1.46}          &   
                        3.81
                        \\ 

Conv-MLP \cite{wang2022conv}  (TIFS 22)        &- &
                        \multicolumn{1}{c}{32.50}        & 
                        \multicolumn{1}{c}{0.90}          & \multicolumn{1}{c}{2.30}          &  
                        7.05
                        \\
FM-ViT \cite{liu2023fm}   (TIFS 23)                     & -&
                        \multicolumn{1}{c}{12.00}          & \multicolumn{1}{c}{0.43}          &  
                        \multicolumn{1}{c}{0.73}       &
                        2.49
                        \\
LPST \cite{li2023learning}   (AAAI 23)            & -&
                        \multicolumn{1}{c}{\textbf{10.0}}          & \multicolumn{1}{c}{0.7}          &  
                        \multicolumn{1}{c}{0.6}       &
                        4.8
                        \\
                        
ViT+AMA+M$^2$\text{A}$^2$\text{E} \cite{yu2024rethinking} (IJCV 24)             &  -& 
                        \multicolumn{1}{c}{17.99}          & \multicolumn{1}{c}{\textbf{0.00}}          &  
                        \multicolumn{1}{c}{\textbf{0.00}}       &
                        3.86
                        \\

FM-CLIP \cite{liu2024fm} (ACMMM 24) &  - & 
                        \multicolumn{1}{c}{11.00}          & \multicolumn{1}{c}{0.38}          &  
                        \multicolumn{1}{c}{0.66}       &
                        \textbf{2.29}

                        \\

CTNet (Ours)                     &  -     & 
                        \multicolumn{1}{c}{13.45}        & 
                        \multicolumn{1}{c}{\textbf{0.00}}          & 
                        \multicolumn{1}{c}{\textbf{0.00}}          &      \textbf{2.29}
                        \\ 

                        
 \bottomrule
\end{tabular} }
\end{table*}

\begin{table}[t]
\centering
\caption{ 
Intra-domain testing on \textbf{CASIA-SURF} \cite{zhang2020casia}  under the fixed-modal scenario.  The evaluation metrics is ACER (\%) $\downarrow$.
} 
\label{tab:SURF-intra}
\begin{tabular}{cccc}
\toprule
Method & APCER $\downarrow$ & BPCER $\downarrow$ & ACER $\downarrow$ \\ \midrule
Conv-MLP \cite{wang2022conv} (TIFS 22)    &  1.5       &  1.8       &   1.6     \\ 
MA-ViT \cite{liu2023ma} (IJCAI 22)&      0.78   &     0.83    &      0.80  \\ 
FM-ViT \cite{liu2023fm} (TIFS 23)&     0.39   &     0.50    &      0.45  \\ 
AADS \cite{deng2023attention} (TIFS 23)     &    0.74     &  0.49       &    0.62    \\
ViT+AMA+M$^2$\text{A}$^2$\text{E} \cite{yu2024rethinking} (IJCV 24) & 0.81 & 0.42 & 0.62 \\
FM-CLIP \cite{liu2024fm} (ACMMM 24) & 0.42 & 0.45 & 0.43 \\
\midrule 
CTNet (Ours)   &   \textbf{ 0.39 }    &  \textbf{0.39 }      &    \textbf{0.39}    \\ \bottomrule
\end{tabular}
\end{table}

\subsection{Experimental comparisons}
 
To evaluate the proposed CTNet, in this Section, we conduct intra- and cross-domain testing under fixed- and missing-modal scenarios, as defined in \cite{yu2023flexible}. The fixed-modal scenario indicates that the test data incorporates all modalities, while the missing-modal scenario represents cases where some modalities, such as IR or depth, may be absent in the test data.

\subsubsection{Intra-domain testing} 
 
In Table \ref{tab:WMCA_intra}, we adopt the fixed-modal scenario to  conduct intra-domain testing results on \textbf{WMCA} \cite{george2019biometric}.  
First, in Table \ref{tab:WMCA_intra}, we observe that the proposed CTNet outperforms previous two-class multi-modal FAS methods across most protocols.
Next, since the spoof attacks in the "Glasses" protocol of Table \ref{tab:WMCA_intra} obscure only small eye regions while preserving most of the real facial regions, we observe that existing FAS methods are significantly influenced by the liveness information extracted from the real facial regions, resulting in relatively poor performance. 
 
In Table~\ref{tab:SURF-intra}, we adopt the fixed-modal scenario to  conduct intra-domain testing results on \textbf{CASIA-SURF} \cite{zhang2020casia}. 
First, since the spoof attacks from \textbf{CASIA-SURF} \cite{zhang2020casia} cover the entire facial region without preserving real face areas, we observe that existing multi-modal methods achieve promising performance in handling these attacks.
In particular, the authors in \cite{liu2024fm} proposed using contrastive language-image pretraining (CLIP) models \cite{radford2021learning} to significantly enhance detection performance.
Next, the proposed CTNet, even without utilizing additional aligned vision and language knowledge from pre-trained CLIP models, still outperforms previous methods and achieves state-of-the-art performance.

\begin{table}
    \centering
    \small
    \caption{ 
    Intra-domain testing on  \textbf{CASIA-SURF} \cite{zhang2020casia}  and \textbf{CASIA-SURF CeFA} \cite{liu2021casia}  under the missing-modal scenario \cite{yu2023flexible}. 
    The evaluation metric is ACER (\%) $\downarrow$. } 
    \label{tab: flexible_intra}
    \begin{tabular}{ccccc}
    \toprule
         \multirow{3}{*}{Method}  & \multicolumn{4}{c}{Protocol : Testing Modalities} \\ \cmidrule{2-5}
         						  & \makecell{P1: \\ RGB} & \makecell{P2: \\ RGB + D} & \makecell{P3: \\ RGB + IR} & \makecell{P4: \\ RGB + D + IR} \\
    
    \midrule
       V-DM \cite{yu2023flexible} (CVPRW 23)  & 7.87 & 2.59 & 9.06 & 4.97 \\
       V \cite{yu2023flexible} (CVPRW 23)  & 10.81 & 1.44 & 26.34 & 3.82 \\
       V-CADM \cite{yu2023flexible} (CVPRW 23) & 6.04 & 3.85 & 5.97 & 3.91 \\
       C-DM  \cite{yu2023flexible} (CVPRW 23) & 36.49 & 7.91 & 35.89 & 11.36 \\
       R50-DM \cite{yu2023flexible} (CVPRW 23) & 14.24 & 9.32 & 18.18 & 8.1 \\
       MMA-FAS \cite{zheng2024towards} (ECCV 24) & 19.86 & 6.96 & 5.34 & 2.85 \\
       MMDG \cite{lin2024suppress} (CVPR 24) &  13.48 &  13.45 & 13.46  &  13.44 \\
    \midrule  
        CTNet (Ours) & \textbf{5.79} & \textbf{1.33} & \textbf{4.71} & \textbf{1.17}  \\
    \bottomrule
    \end{tabular}
\end{table}
 
In Table~\ref{tab: flexible_intra},  we adopt the  missing-modal scenario introduced in \cite{yu2023flexible} to  conduct intra-domain testing results on  \textbf{CASIA-SURF} \cite{zhang2020casia}  and \textbf{CASIA-SURF CeFA} \cite{liu2021casia}. 
First, by comparing the cases of P1: RGB, P2: RGB+D, P3: RGB+IR, and P4: RGB+D+IR, we observe that missing modalities cause performance degradation, especially in the case where depth modalities are missing (i.e.,  P1: RGB and P3: RGB+IR).
Next, by comparing the cases of RGB+D vs. RGB+IR, the results show that fine-grained depth information may be more effective in facilitating the detection of spoof attacks compared to IR information.
Finally, the proposed CTNet, by extracting IR-like and depth-like features from the proposed the proposed complementary feature learning as replacements, effectively tackles the issue of missing modalities.

\begin{table}[t]
\centering
\caption{ Cross-domain testing on
 \textbf{WMCA} \cite{george2019biometric}, \textbf{CASIA-SURF} \cite{zhang2020casia}, and \textbf{CASIA-SURF CeFA} \cite{liu2021casia} under the fixed-modal scenario. The evaluation metric is ACER (\%) $\downarrow$.} 
\label{tab:inter-domain}
\resizebox{1\columnwidth}{!}{%
\begin{tabular}{cccc}
\toprule
Method & \makecell{\textbf{CASIA-SURF CeFA} \\ $\rightarrow$ \textbf{CASIA-SURF}} & \makecell{\textbf{CASIA-SURF } \\ $\rightarrow$ \textbf{CASIA-SURF CeFA}}& \makecell{\textbf{WMCA} \\ $\rightarrow$ \textbf{CASIA-SURF}} \\
\midrule
MA-ViT \cite{liu2023ma} (IJCAI 22) & - & - & 10.41 \\
FM-ViT \cite{liu2023fm} (TIFS 23) & 10.24 & 21.43$\pm$1.24 & 8.45 \\
ViT+AMA+M$^2$\text{A}$^2$\text{E} \cite{yu2024rethinking} (IJCV 24) & - & - & 8.60 \\
MMDG \cite{lin2024suppress} (CVPR 24) & 7.59 & 15.57$\pm$1.15 & 7.42 \\
\midrule  
CTNet (Ours) & \textbf{5.54} & \textbf{8.9 $\pm$ 3.05} & \textbf{5.00} \\
\midrule
Method &\makecell{\textbf{WMCA} \\ $\rightarrow$ \textbf{CASIA-SURF CeFA}}& \makecell{\textbf{CASIA-SURF CeFA} \\ $\rightarrow$ \textbf{WMCA} } & \makecell{\textbf{CASIA-SURF} \\ $\rightarrow$ \textbf{WMCA} }\\
\midrule
MA-ViT \cite{liu2023ma} (IJCAI 22) & - & - & 20.63 \\
FM-ViT \cite{liu2023fm} (TIFS 23) & 8.77$\pm$3.13 & 26.34 & 20.00 \\
ViT+AMA+M$^2$\text{A}$^2$\text{E} \cite{yu2024rethinking} (IJCV 24) & - & - & 18.83 \\
MMDG \cite{lin2024suppress} (CVPR 24) & 11.68$\pm$3.62 & 29.57 & 30.00 \\
\midrule 
CTNet (Ours) & \textbf{6.07$\pm$2.97} & \textbf{24.31} & \textbf{14.96} 
\\
\bottomrule
\end{tabular}
}
\end{table}

\subsubsection{Cross-domain testing} 
 
In Table~\ref{tab:inter-domain}, we adopt the fixed-modal scenario to conduct cross-domain testing results on \textbf{WMCA} \cite{george2019biometric}, \textbf{CASIA-SURF} \cite{zhang2020casia}, and \textbf{CASIA-SURF CeFA} \cite{liu2021casia} by using only one dataset for training and the remaining two datasets for testing.
First, since \textbf{WMCA} and \textbf{CASIA-SURF CeFA}, collected using the same camera and data preprocessing, exhibit small distribution discrepancies, we observe that existing multi-modal FAS methods achieve promising performance in the protocols \textbf{WMCA} $\rightarrow$ \textbf{CASIA-SURF CeFA} and \textbf{CASIA-SURF CeFA} $\rightarrow$ \textbf{WMCA}.
Next, because \textbf{WMCA} covers a broader range of attack types and includes subjects with more diverse skin tones compared to \textbf{CASIA-SURF CeFA} and \textbf{CASIA-SURF}, we observe that existing multi-modal FAS methods achieve promising performance on the protocols \textbf{WMCA} $\rightarrow$ \textbf{CASIA-SURF CeFA} and \textbf{WMCA} $\rightarrow$ \textbf{CASIA-SURF}. However, their performance significantly drops on the protocols \textbf{CASIA-SURF CeFA} $\rightarrow$ \textbf{WMCA} and \textbf{CASIA-SURF} $\rightarrow$ \textbf{WMCA}.
By learning consistent feature transitions from live samples and inconsistent feature transitions between live and spoof samples, the proposed CTNet outperforms existing multi-modal FAS methods and achieves state-of-the-art performance.
Notably, it demonstrates superior domain generalization ability by excelling in challenging protocols such as \textbf{CASIA-SURF CeFA} $\rightarrow$ \textbf{WMCA} and \textbf{CASIA-SURF} $\rightarrow$ \textbf{WMCA}, which are characterized by large distribution discrepancies between seen and unseen spoof attacks, thereby improving the detection of unknown spoof attacks.

\begin{table}
    \small
    \centering
    \caption{ Cross-domain testing on [ \textbf{CASIA-SURF CeFA}, \textbf{CASIA-SURF}] $\rightarrow$ \textbf{WMCA} under the missing-modal scenario \cite{yu2023flexible}. 
    The evaluation metric is ACER $\downarrow$. 
    }
    \label{tab: flexible_inter}
    \begin{tabular}{ccccc}
    \toprule
         \multirow{3}{*}{Method}  & \multicolumn{4}{c}{Protocol / Testing Modalities} \\ \cmidrule{2-5}
         						  & \makecell{P1: \\ RGB} & \makecell{P2: \\ RGB + D} & \makecell{P3: \\ RGB + IR} & \makecell{P4: \\ RGB + D  + IR} \\
    \midrule
        V-DM \cite{yu2023flexible} (CVPRW 23)  & 39.06 & 30.37 & 40.61 & 29.51 \\
       V \cite{yu2023flexible} (CVPRW 23)  & 48.52 & 44.42 & 50.00 & 33.08 \\
       V-CADM \cite{yu2023flexible} (CVPRW 23) & 42.38 & 33.87 & 37.81 & 33.59 \\
       C-DM  \cite{yu2023flexible} (CVPRW 23) & 32.25 & 25.44 & 32.95 & 26.92 \\
       R50-DM \cite{yu2023flexible} (CVPRW 23) & 46.45 & 25.36 & 46.53 & 19.43 \\
       MMA-FAS \cite{zheng2024towards} (ECCV 24) & 34.07 & 30.49 & 31.58 & 29.08 \\
       MMDG \cite{lin2024suppress} (CVPR 24) & 23.04  &  19.91 &  20.66 &  16.80 \\
    \midrule  
        CTNet (Ours) & \textbf{22.56} & \textbf{17.16} & \textbf{18.76} & \textbf{16.51}  \\
    \bottomrule
    \end{tabular}
\end{table}
 
In Table~\ref{tab: flexible_inter}, we adopt the fixed-modal scenario to conduct cross-domain testing results on the protocols [\textbf{CASIA-SURF CeFA}, \textbf{CASIA-SURF}] $\rightarrow$ \textbf{WMCA}. 
Recall that \textbf{WMCA} covers a broader range of attack types and includes subjects with more diverse skin tones compared to \textbf{CASIA-SURF CeFA} and \textbf{CASIA-SURF}. Consequently, the protocols [\textbf{CASIA-SURF CeFA}, \textbf{CASIA-SURF}] $\rightarrow$ \textbf{WMCA} become doubly challenging when encountering unseen spoof attacks and missing modalities.
Hence, we see that previous multi-modal FAS methods still exhibit a large performance gap, as shown in Table~\ref{tab: flexible_inter}.
By incorporating complementary feature learning alongside consistent and inconsistent feature transition learning, the proposed CTNet effectively tackles unseen attacks, even in the missing-modal scenario and when encountering overlapping unseen spoof attacks.

\section{Conclusion}
 
In this paper, we propose a novel Cross-modal Transition-guided Network (CTNet) to address the multi-modal face anti-spoofing (FAS) problem.
First, we investigate the characteristics of the consistent feature transitions from live faces and the inconsistent feature transitions between live and spoof faces.
Next, we propose a novel cross-modal transition-guided feature learning  to learn the above characteristics within the multi-modal FAS task for facilitating the detection of out-of-distribution (OOD) attacks during inference.
Furthermore, we propose an effective complementary feature learning to learn the IR-like and depth-like features from RGB images  as replacements  when encountering the missing modalities.
Extensive experiments show that the proposed CTNet outperforms previous two-class multi-modal FAS methods across most protocols. 
 


\bibliographystyle{elsarticle-num-names} 
\bibliography{reference}
\end{document}